\documentclass[letterpaper, 10 pt, conference]{ieeeconf}
\usepackage{times}
\usepackage[pdftex]{graphicx}
\usepackage{subfigure}
\usepackage{amsmath,amssymb,amsopn,amstext,amsfonts}
\usepackage{cancel}
\usepackage[space]{cite}
\usepackage{pdfsync}
\usepackage{balance}
\usepackage{color}
\usepackage{mathtools}
\usepackage{makecell}
\usepackage{gensymb}
\usepackage[ruled,vlined]{algorithm2e}
\usepackage{bm}
\usepackage{diagbox}
\usepackage{float}
\usepackage{epstopdf}
\usepackage{pifont}
\usepackage{multirow}
\usepackage{url}
\usepackage{tabularx}
\usepackage{booktabs}
\usepackage{bm}
\usepackage{multirow}
\usepackage{threeparttable}

\usepackage{xcolor}
\usepackage[linkcolor=black,citecolor=black,urlcolor=black,colorlinks=true]{hyperref}

\usepackage{verbatim} 

\newtheorem{problem}{Problem}[section]

\graphicspath{{../figure/}}
\DeclareGraphicsExtensions{.png,.jpg,.eps,.pdf}
\IEEEoverridecommandlockouts
\title{\LARGE \bf Simultaneous Time Synchronization and \\
Mutual Localization for Multi-robot System}
\author{Xiangyong Wen\textsuperscript{1,4,*}, Yingjian Wang\textsuperscript{2,4,*}, Xi Zheng\textsuperscript{5}, Kaiwei Wang\textsuperscript{3}, Chao Xu\textsuperscript{2,4}, Fei Gao\textsuperscript{2,4}
\thanks{\textsuperscript{1}Polytechnic Institute of Zhejiang University, Hangzhou, 310027, China.}
\thanks{\textsuperscript{2}Institute of Cyber-Systems and Control, Zhejiang University, Hangzhou, 310027, China.}
\thanks{\textsuperscript{3}Collegue of Optical Science and Engineering, Zhejiang University, Hangzhou, 310027, China.}
	\thanks{\textsuperscript{4}Huzhou Institute of Zhejiang University, Huzhou, 313000, China.}
    \thanks{\textsuperscript{5}The Hong Kong Polytechnic University, Hong Kong, 100872, China.}
	\thanks{Corresponding author: Fei Gao and Yingjian Wang. E-mails:{
			\tt\small 
			\{wenxiangyong, yj\_wang, fgaoaa\}@zju.edu.cn.
		}*Equal contributors.}
  }
\begin{document}

\maketitle
\thispagestyle{empty}
\pagestyle{empty}

\begin{abstract}
 Mutual localization stands as a foundational component within various domains of multi-robot systems.
 Nevertheless, in relative pose estimation,  time synchronization is usually underappreciated and rarely addressed, although it significantly influences estimation accuracy.
 In this paper, we introduce time synchronization into mutual localization to recover the time offset and relative poses between robots simultaneously.
 Under a constant velocity assumption in a short time, we fuse time offset estimation with our previous bearing-based mutual localization by a novel error representation.
 Based on the error model, we formulate a joint optimization problem and utilize semi-definite relaxation (SDR) to furnish a lossless relaxation.
 By solving the relaxed problem, time synchronization and relative pose estimation can be achieved when time drift between robots is limited. 
 To enhance the application range of time offset estimation, we further propose an iterative method to recover the time offset from coarse to fine.
 Comparisons between the proposed method and the existing ones through extensive simulation tests present prominent benefits of time synchronization on mutual localization.
 Moreover, real-world experiments are conducted to show the practicality and robustness.
\end{abstract}

\section{Introduction}
\label{sec:introduction}
Multi-robot systems have gained more and more attention from industry and academia due to their strong environmental adaptability, efficient mission completion, and high fault tolerance in complex tasks, such as multi-robot exploration\cite{gao2022} and post-disaster rescue \cite{sugiyama2008coordination}. 
In the collaborative tasks of multi-robot systems, sharing coordinate frames plays a crucial role in providing localization consensus. 
Some works rely on external devices, such as GPS, motion capture, and UWB, to achieve coordinate frame sharing.
The dependence on external devices limits the applications of these multi-robot systems in many real-world environments, such as denied or sheltered scenarios. 
Therefore, performing relative localization with only onboard sensors has recently become a focus of multi-robot research. 

Among various relative localization approaches, 
methods \cite{zhou2012determining, franchi2013mutual, nguyen2020vision} that leverage mutual observation measurements between robots to recover relative poses have attracted much attention since they are independent of environment and require quite a low communication bandwidth.
In our previous work \cite{wang2022certifiably}, relative pose estimation with optimality guarantees was achieved by utilizing bearing measurements between robots. 
However, many drawbacks, such as time synchronization, still limit the practical application of relative localization methods.
It denotes that data from different robots have different timestamps, typically from the unsynchronized clocks and sensors triggering delay.
Such time offsets may lead to incorrect data associations, divergence in relative pose estimation, and even potential collapse of multi-robot systems.
Some efforts have attempted to address this issue using network protocols like Network Time Protocol (NTP)\cite{mills2010network} and Precision Time Protocol (PTP)\cite{eidson2002ieee}.
These methods usually require a certain hardware and may have poor performance with high network latency.
Besides, time synchronization has been extensively studied in multi-sensor fusion, where researchers aim to calibrate the time offset between sensors.
For example, Qin \cite{qin2018online} introduced time offset estimation between cameras and IMUs in visual-inertial odometry, improving estimation accuracy.

\begin{figure}[t]
	\centering
	\includegraphics[width=0.5\textwidth]{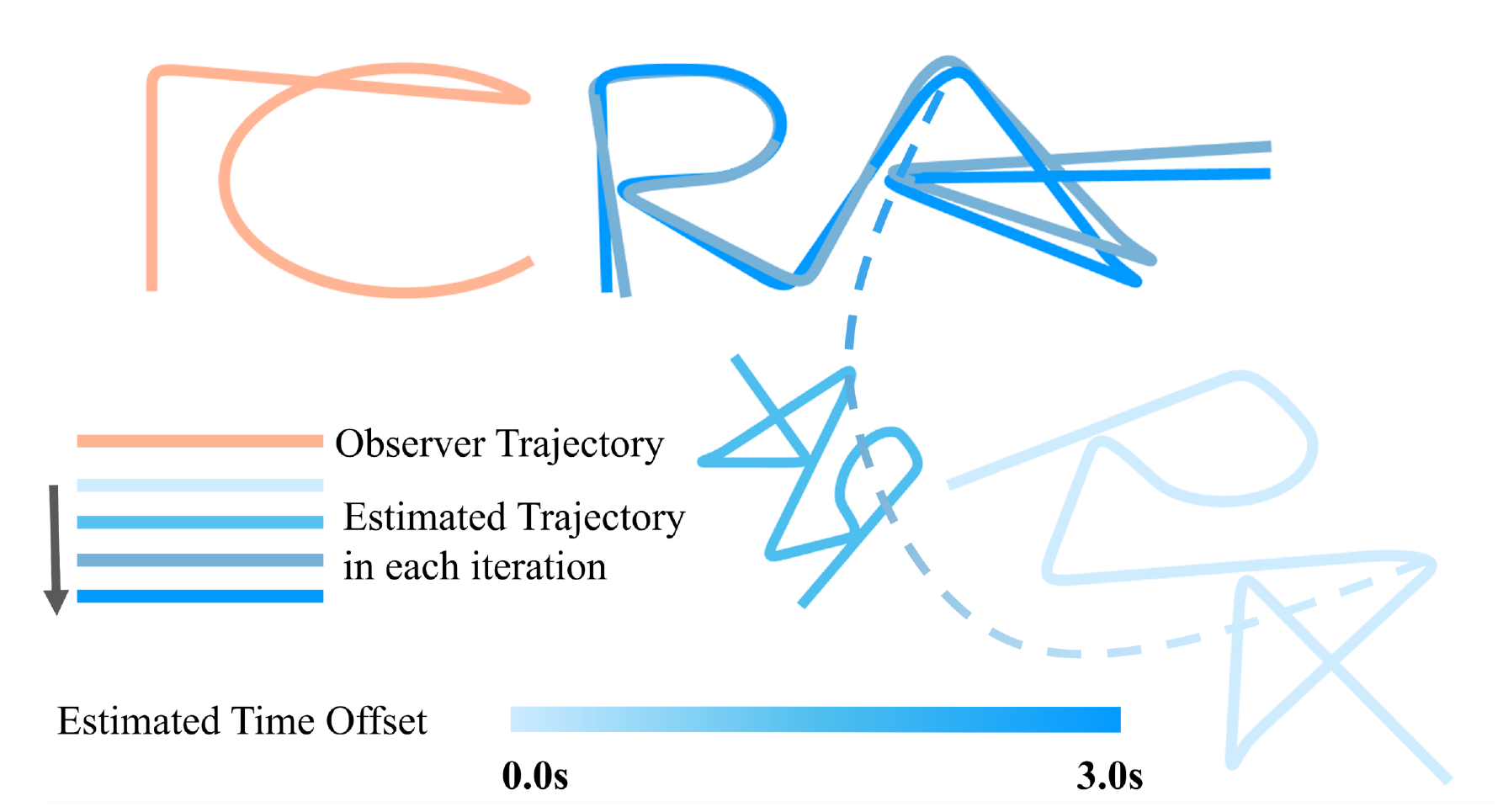}
	\caption{\label{fig: head}The ‘ICRA’ trajectory recovering in an extremely large time offset situation. It can be seen that the trajectory of the letter 'RA' gradually approaches the ground truth 'IC' frame as the number of iterative optimizations increases.}
	\vspace{-1cm}
\end{figure}

Inspired by Qin \cite{qin2018online}, this paper first introduces the online time offset estimation into mutual localization to solve time synchronization and relative pose estimation simultaneously. 
In most cases, the robot's velocity can be seen as constant if time is short. Based on this, we introduce time offset into the observation model, resulting in a novel error representation, including time offset and relative pose.
Since the error representation can be reformulated into linear form, the joint state estimation can be formulated as a quadratically constrained quadratic problem (QCQP) by introducing auxiliary variables. 
Then, this problem is further relaxed into a convex programming by semi-definite relaxation, through which a globally optimal solution can be obtained in polynomial time.
Furthermore, considering that the constant velocity assumption will be invalid when the time offset is large, we design an iterative optimization framework, which recovers the time offset and relative pose from coarse to fine. 
Extensive simulation and real-world experiments verify our proposed method's outperforming accuracy and robustness in multi-robot localization scenarios.

In summary, our contributions are listed as follows:
\begin{enumerate}
    \item We introduce time offset estimation into multi-robot relative localization, deriving a novel error representation and joint estimation problem.
    \item We design an iterative optimization strategy, significantly enhancing the accuracy and range of time offset estimation. 
    \item We conduct extensive simulations and real-world experiments to validate the feasibility of the proposed method.  
\end{enumerate}


\section{Related works}
\label{sec:related_works}

\subsection{Mutual Localization}
\label{Time_synchronization}
The mainstream of mutual localization can be categorized into map-based methods and mutual observation-based methods. 
Map-based methods \cite{howard2006multi, melnyk2012cooperative, dong2015distributed, lajoie2020door} take the environment features as the basis and realize mutual localization by environment landmarks sharing and loop closure detection.
However, this approach occupies a large computational resource and is significantly affected by the surrounding environment. 
On the contrary, mutual observation-based approaches \cite{zhou2012determining, franchi2013mutual, nguyen2020vision} focus only on relative measurements between robots, such as distance and bearing between robots.
Stegagno \cite{stegagno2011mutual} proposed a probabilistic multiple formation registration method for anonymous bearing-only measurements localization based on the particle filter. 
Zhou \cite{zhou2012determining} summarized the mutual localization question into 14 minimal problems by the symmetry of range and bearing measurements and provided algebraic and numerical solutions to solve these minimal problems.
Our previous work \cite{wang2022certifiably} formulated the mutual localization as a mixed integer quadratically constrained quadratic problem (MIQCQP) and relaxed this nonconvex problem into an SDP problem to get a certifiably global optimal solution.
As far as we know, existing works did not discuss the time synchronization issue in mutual localization, although the time drift between robots will severely influence the performance of multi-robot systems.

\subsection{Time synchronization}
Researchers usually use network protocols to solve the time synchronization problem in multi-robot systems, such as NTP \cite{mills2010network} and PTP \cite{eidson2002ieee}.
Among that, NTP has a theoretical accuracy of 1 ms and does not rely on additional hardware. 
However, its accuracy is severely affected by network latency and bandwidth limitations.
In multi-robot systems, frequent communication between robots may lead to discontinuous data streams and incorrect time offset estimation. 
The accuracy of PTP is up to 1us but needs specialized and expensive hardware equipment. 
The limitation of weight and price restricts its application, especially for low-cost multi-robot systems.
On the other hand, researchers in communication networks considered system time offset as an unknown parameter and used estimation algorithms to solve time synchronization problems.
Yang \cite{yang2010adaptive} employed an interactive multi-model Kalman filter that adaptively calculated clock skew based on environmental changes.
In Shi's method\cite{shi2019novel}, a maximum likelihood estimation with an innovative implementation method was presented to minimize the time delay.  
Like NTP, these methods rely on stable connections between agents, limiting the application in multi-robot systems. 

In addition, multi-sensor fusion for individual robots faces a similar time synchronization problem. 
Furgale \cite{furgale2013unified} released a Kalibr toolbox, widely used to calibrate the time offset and extrinsic parameters between the camera and IMU by continuous-time batch estimation. 
Qin \cite{qin2018online} integrated time offset calibration into the simultaneous localization and mapping (SLAM) optimization framework and solved it in parallel with the robot state estimation problem.
Inspired by the above works, we utilize robots' motion in this paper to recover the time offset and estimate the relative poses simultaneously. 
The tightly coupled optimization brings high accuracy and robustness in practice. 

\section{Relative Localization with Time Offset Estimation}
\label{sec: problem formulation}
We study the bearing-based relative localization, which uses only local odometry and inter-robot bearing measurements to recover the relative poses.
In this section, we introduce time synchronization into relative localization to optimize the offset and relative pose simultaneously. 
First, we utilize uniform velocity assumption to derive a novel error format.
Thanks to the new error representation, the time offset between trajectories can be optimized continuously. 
Furthermore, exploiting the linear form of error, we construct a least square problem.
Then, by Shur Complement, we obtain a marginalized problem to recover time offset and relative rotations.

\subsection{Novel Error Representation with Time Offset}
\label{error_representation}
\begin{figure}[t]
	\centering
	\includegraphics[width=0.5\textwidth]{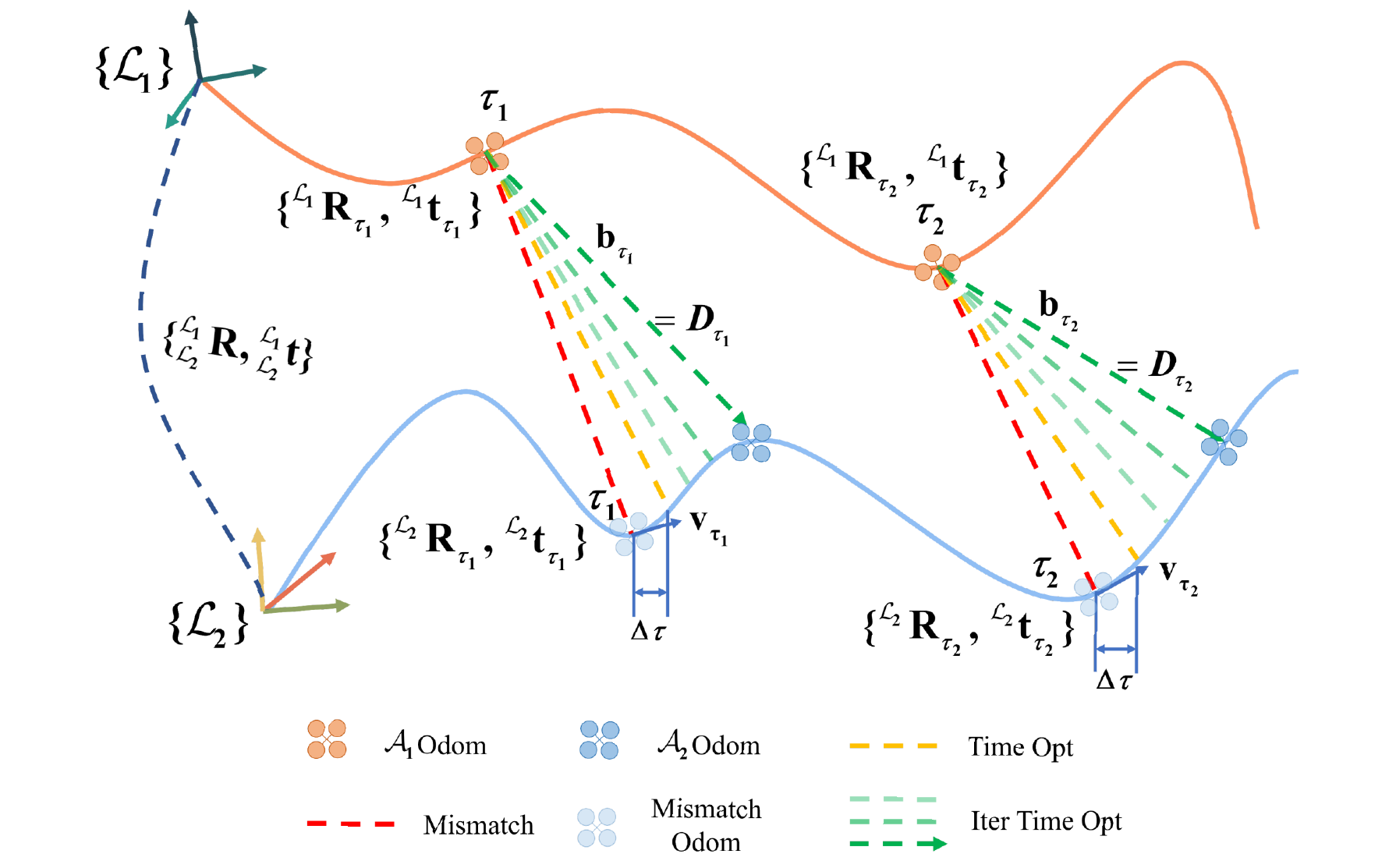}
	\caption{\label{fig: method} Demonstration of the time offset estimation and mutual localization in bearing-based multi-robot system. }
	\vspace{-0.7cm}
\end{figure}

We consider two robots: a observing robot $\mathcal{A}_{1}$ and a observed robot $\mathcal{A}_{2}$.
As shown in Fig.\ref{fig: method}, the corresponding local odoemtry frame are defined as $\mathcal{L}_{1}$ and $\mathcal{L}_{2}$.
$^{\mathcal{L}_{n}}\mathbf{t}_{\tau_{k}} \in \mathbb{R}^3$ and $^{\mathcal{L}_{n}}\mathbf{R}_{\tau_{k}} \in SO(3)$ denotes the translation and rotation in frame $\mathcal{L}_{n}$ at $\mathcal{A}_{1}$'s local time $\tau_{k}$.
 We emphasize the definition of local time, since if there is a time offset $\Delta \tau$ between robots, each robot will have a different local time even at the same "world" time.
 Similarly, $^{\mathcal{A}_{1}}\mathbf{b}^{\mathcal{A}_{2}}_{\tau_{k}}$ denotes the bearing measurement of  $\mathcal{A}_{2}$ in the frame of $\mathcal{A}_{1}$ at $\mathcal{A}_{1}$' local time $\tau_{k}$.
 Our goal is to estimate the relative frame transformation, ${ }_{\mathcal{L}_{2}}^{\mathcal{L}_{1}}\mathbf{t}$ and ${ }_{\mathcal{L}_{2}}^{\mathcal{L}_{1}}\mathbf{R}$, and time offset $\Delta \tau$, using each robot's local odometry and inter-robot bearing measurements.

In our previous work, the estimation error term can be written as: 
\begin{equation}
\label{equ:instance0}
 \mathbf{e}_{\tau_{k}} = {}^{\mathcal{L}_{1}}\mathbf{R}_{\tau_{k}}\mathbf{b}_{\tau_{k}}D_{\tau_{k}}+{}^{\mathcal{L}_{1}}\mathbf{t}_{\tau_{k}} - ({ }_{\mathcal{L}_{2}}^{\mathcal{L}_{1}}\mathbf{R} {}^{\mathcal{L}_{2}}\mathbf{t}_{\tau_{k}}+{ }_{\mathcal{L}_{2}}^{\mathcal{L}_{1}}\mathbf{t}),
\end{equation}
where ${ }_{\mathcal{L}_{2}}^{\mathcal{L}_{1}}\mathbf{R}$, ${ }_{\mathcal{L}_{2}}^{\mathcal{L}_{1}}\mathbf{t}$ and inter-robot distance $D_{\tau_{k}}$ are unknown variables to be solved.
When the local time of two robots has been synchronized, the error makes sense.
However, if there is any time offset $\Delta \tau$ between $\mathcal{A}_{1}$ and $\mathcal{A}_{2}$,
this error modeling is not precise anymore.
That's because when $\mathcal{A}_{1}$ is with state ${}^{\mathcal{L}_{1}}\mathbf{R}_{\tau_{k}}$ and ${}^{\mathcal{L}_{1}}\mathbf{t}_{\tau_{k}}$, the position of $\mathcal{A}_{2}$ corresponding to the bearing 
$^{\mathcal{A}_{1}}\mathbf{b}^{\mathcal{A}_{2}}_{\tau_{k}}$ is not ${}^{\mathcal{L}_{2}}\mathbf{t}_{\tau_{k}}$ but ${}^{\mathcal{L}_{2}}\mathbf{t}_{(\tau_{k} + \Delta \tau)}$.
Thus, the estimation error with time offset should be correlated as 
\begin{equation}
\label{equ:instance0}
 \mathbf{e}_{\tau_{k}} = {}^{\mathcal{L}_{1}}\mathbf{R}_{\tau_{k}}\mathbf{b}_{\tau_{k}}D_{\tau_{k}}+{}^{\mathcal{L}_{1}}\mathbf{t}_{\tau_{k}} - ({ }_{\mathcal{L}_{2}}^{\mathcal{L}_{1}}\mathbf{R} {}^{\mathcal{L}_{2}}\mathbf{t}_{(\tau_{k} + \Delta \tau)}+{ }_{\mathcal{L}_{2}}^{\mathcal{L}_{1}}\mathbf{t}).
\end{equation}

The correlation is simple but very significant to relative pose estimation.
If the time offset is not considered, the bias will exist throughout the entire estimation process and affect the accuracy, which is verified in the subsequent experiments in Sec.\ref{sec:experiment}.
Then, to optimize the time offset, it is necessary to model ${}^{\mathcal{L}_{2}}\mathbf{t}_{(\tau_{k} + \Delta \tau)}$ using $\Delta \tau$ precisely. 
Here, we borrow the idea that the pixel has constant 2D velocity on the image in sensor time calibration \cite{qin2018online}.  
We consider that robots move with a constant velocity in a short time and thus obtain the approximation of ${}^{\mathcal{L}_{2}}\mathbf{t}_{(\tau_{k} + \Delta \tau)}$ as 
\begin{equation}
    {}^{\mathcal{L}_{2}}\mathbf{t}_{(\tau_{k} + \Delta \tau)} \approx {}^{\mathcal{L}_{2}}\mathbf{t}_{\tau_{k}} + {}^{\mathcal{L}_{2}}\mathbf{v}_{\tau_{k}} \Delta \tau,
\end{equation}
where ${}^{\mathcal{L}_{2}}\mathbf{v}_{\tau_{k}}$ is the velocity of robot $\mathcal{A}_{2}$, which can be obtained by local odometry.
This constant velocity assumption is reliable since actual robots are always desired to move smoothly in navigation. 
In this way, we effectively introduce the time offset $\Delta \tau$ into the relative pose estimation and derive a novel error representation:
\begin{equation}
\label{equ:instance1}
\begin{aligned}
    \mathbf{e}_{\tau_{k}} =& ^{\mathcal{L}_{1}}\mathbf{R}_{\tau_{k}}\mathbf{b}_{\tau_{k}}D_{\tau_{k}}+ {}^{\mathcal{L}_{1}}\mathbf{t}_{\tau_{k}} - \\
    &({}_{\mathcal{L}_{2}}^{\mathcal{L}_{1}}\mathbf{R} ^{\mathcal{L}_{2}}\mathbf{t}_{\tau_{k}} + {}_{\mathcal{L}_{2}}^{\mathcal{L}_{1}}\mathbf{R} ^{\mathcal{L}_{2}}\mathbf{v}_{\tau_{k}} \Delta \tau + { }_{\mathcal{L}_{2}}^{\mathcal{L}_{1}}\mathbf{t}).
\end{aligned}
\end{equation}
The new error modeling jointly considers time offset and relative poses with the support of velocity estimation.
It fully utilizes the robot's motion information.

\subsection{Estimation Problem Formulation}
We then construct a joint optimization program using the aforementioned error representation.
Observing that $\Delta \tau$ is unknown, we consider to take ${}_{\mathcal{L}_{2}}^{\mathcal{L}_{1}}\mathbf{R}$ and $\Delta\tau {}_{\mathcal{L}_{2}}^{\mathcal{L}_{1}}\mathbf{R}$ as two independent variables.
Thus, we introduce
\begin{equation}
\label{equ:instance4}
\begin{aligned}
    \mathbf{r}_p & = \text{vec}({ }_{\mathcal{L}_{2}}^{\mathcal{L}_{1}}\mathbf{R}) \in \mathbb{R}^{9}, \\
    \mathbf{r}_s & = \text{vec}(\Delta\tau{ }_{\mathcal{L}_{2}}^{\mathcal{L}_{1}}\mathbf{R}) \in \mathbb{R}^{9}.
\end{aligned}
\end{equation}
Since $\mathbf{Rx}=(\mathbf{x}^T\otimes\mathbf{I_{3}})\text{vec}(\mathbf{R})$, where $\otimes$ denotes Kronecker product,
the error (\ref{equ:instance1}) can be rewritten in linear form  
\begin{equation}	
\label{equ:instance3}
\begin{aligned}
    \mathbf{e}_{\tau_{k}} =\ &\mathbf{g}_{\tau_{k}} D_{\tau_{k}} + {}^{\mathcal{L}_{1}}\mathbf{t}_{\tau_{k}} - {}_{\mathcal{L}_{2}}^{\mathcal{L}_{1}}\mathbf{t} \\
    &-(^{\mathcal{L}_{2}}\mathbf{t^T_{\tau_{k}}}\otimes\mathbf{I_{3}})\mathbf{r_p} - (^{\mathcal{L}_{2}}\mathbf{v^T_{\tau_{k}}}\otimes\mathbf{I_{3}})\mathbf{r_s},
\end{aligned}
\end{equation}
where $\mathbf{g}_{\tau_{k}}={}^{\mathcal{L}_{1}}\mathbf{R}_{\tau_{k}}\mathbf{b}_{\tau_{k}}$.
Then, we introduce a homogenizing variable $y$ with $y^2 = 1$ to rid a quadratic equation of linear terms.  
Defining a complete optimization variable $\mathbf{x}$ and data matrix $\mathbf{A}_{\tau_k}$, we obtain  $\mathbf{e}_{\tau_k} = \mathbf{A}_k \mathbf{x}$ where
\begin{equation}	
\label{equ:instance3}
\begin{aligned}
    \mathbf{x} = [&\ \mathbf{r}_s^T,\ \mathbf{r}_p^T, \ y,\ {}_{\mathcal{L}_{2}}^{\mathcal{L}_{1}}\mathbf{t}, \ D_{\tau_{1}}, \cdots, D_{\tau_{N}}]^T \in \mathbb{R}^{(22+N)}, \\
    \mathbf{A}_{\tau_k} = [&-{}^{\mathcal{L}_{2}}\mathbf{t}^T_{\tau_k}\otimes\mathbf{I}_3, -{}^{\mathcal{L}_2}\mathbf{v}^T_{\tau_k}\otimes\mathbf{I}_3, {}^{\mathcal{L}_{1}}\mathbf{t}_{\tau_{k}}, -\mathbf{I}_{3}, \\ 
    & \mathbf{0}_{3\times 1}, \cdots,\mathbf{g}_{\tau_k},  \cdots \mathbf{0}_{3\times 1}]
\end{aligned}
\end{equation}
where $N$ is the number of total measurements.
Subsequently, we can derive a least square problem: 
\begin{equation}
\label{equ:instance9}
\begin{aligned}
    \mathbf{x}^* &=\underset{\mathbf{x}}{\arg\min}\ \sum_{k=1}^{N}(\mathbf{e}_{\tau_k})^T \mathbf{W}_{\tau_k} \mathbf{e}_{\tau_k}\\
    &=\underset{\mathbf{x}}{\arg\min}\ \mathbf{x}^T \mathbf{Q} \mathbf{x}\\
    s.t.& \ \ \   y^2=1,\ { }_{\mathcal{L}_{2}}^{\mathcal{L}_{1}}\mathbf{R} \in SO(3),
\end{aligned}
\end{equation}
where $\mathbf{W}_{\tau_k}$ is the weight matrix and the data matrix $\mathbf{Q} = \sum_{k=1}^{N} \mathbf{A}_{\tau_k}^T \mathbf{W}_{\tau_k} \mathbf{A}_{\tau_k}$.
Then, we eliminate the unconstrained distance and relative translation variables.
Define the truncted optimization variable $\widetilde{\mathbf{x}} = [\ \mathbf{r}^T_s,\ \mathbf{r}^T_p,\ y]^T$ and divide $\mathbf{Q}$ into four blocks 
\begin{gather}
	\mathbf{Q} = \begin{bmatrix} \mathbf{A} & \mathbf{B} \\
	\mathbf{B}^T & \mathbf{C} \end{bmatrix},
\end{gather}
where $\mathbf{A}$ represent the block corresponding to $\widetilde{\mathbf{x}}$. 
The unconstrained variables are marginalized by Schur Complement, and we obtain
\begin{gather}
\label{pro:marg0}
    \widetilde{\mathbf{x}}^* =\underset{\widetilde{\mathbf{x}}}{\arg\min}\ \widetilde{\mathbf{x}}^T \bar{\mathbf{Q}} \widetilde{\mathbf{x}} \notag \\
    s.t.\ y^2=1,\ { }_{\mathcal{L}_{2}}^{\mathcal{L}_{1}}\mathbf{R} \in SO(3), \notag
\end{gather}
where $\bar{\mathbf{Q}} = \mathbf{A} -  \mathbf{B}  \mathbf{C}^{-1} \mathbf{B}^T$.
In this way, the optimization variable only involves the norm-1 and rotation constraints.

\section{Relaxation and Iterative Optimization}\label{Iter_Op}
\begin{table*}[t] 
\label{tab:cost}
\centering
\caption{Constraints and Constraint Matrix}
\begin{threeparttable}
    \begin{tabular}{cccc}
    \hline
    \textbf{Constraint}&
    \multicolumn{1}{c}{
        \begin{tabular}[c]{@{}c@{}}
            \\$(n{}_{\mathcal{L}_{2}}^{\mathcal{L}_{1}}\mathbf{R})^{\mathbf{T}} (n{}_{\mathcal{L}_{2}}^{\mathcal{L}_{1}}\mathbf{R}) = n^2\mathbf{I}_3$\\ \\
    \end{tabular}} &  
    $(n{}_{\mathcal{L}_{2}}^{\mathcal{L}_{1}}\mathbf{R}) (n{}_{\mathcal{L}_{2}}^{\mathcal{L}_{1}}\mathbf{R})^{\mathbf{T}} =  n^2\mathbf{I}_3$ &
    $(n{}_{\mathcal{L}_{2}}^{\mathcal{L}_{1}}\mathbf{R})^{(i)} \times (n{}_{\mathcal{L}_{2}}^{\mathcal{L}_{1}}\mathbf{R})^{(j)} =n(n{}_{\mathcal{L}_{2}}^{\mathcal{L}_{1}}\mathbf{R})^{(k)}$ \\
    \hline
    $\mathbf{Q}_i$ &
    \multicolumn{1}{c}{
        \begin{tabular}[c]{@{}c@{}}
					\\$\left[\begin{array}{cccc}
			\mathbf{0}_{9\times{9}} & \mathbf{0}_{9\times{9}} & \mathbf{0}_{9\times{1}} & \mathbf{0}_{9\times{1}} \\
			\mathbf{0}_{9\times{9}} & \mathbf{J}_3 \otimes \mathbf{I}_3 & \mathbf{0}_{9\times{1}} & \mathbf{0}_{9\times{1}} \\
			\mathbf{0_{1\times{9}}} & \mathbf{0_{1\times{9}}} &  \alpha & 0 \\
			\mathbf{0_{1\times{9}}} & \mathbf{0_{1\times{9}}} & 0 &\beta
        \end{array}\right]$\\ \\
    \end{tabular}}
     &  
  $\left[\begin{array}{cccc}
			\mathbf{0}_{9\times{9}} & \mathbf{0}_{9\times{9}} & \mathbf{0}_{9\times{1}} & \mathbf{0}_{9\times{1}} \\
			\mathbf{0}_{9\times{9}} & \mathbf{I_{3}} \otimes \mathbf{J_{3}} & \mathbf{0}_{9\times{1}} & \mathbf{0}_{9\times{1}} \\
			\mathbf{0_{1\times{9}}} & \mathbf{0_{1\times{9}}} & \alpha & 0 \\
			\mathbf{0_{1\times{9}}} & \mathbf{0_{1\times{9}}} & 0 &\beta
		\end{array}\right]$ &
  $\left[\begin{array}{cccc}
			\mathbf{U} & \mathbf{0}_{9\times{9}} & \mathbf{0}_{9\times{1}} & \mathbf{0}_{9\times{1}} \\
			\mathbf{0}_{9\times{9}} & \mathbf{V} & \mathbf{u} & \mathbf{v} \\
			\mathbf{0_{1\times{9}}} & \mathbf{0_{1\times{9}}} & 0 & 0 \\
			\mathbf{0_{1\times{9}}} & \mathbf{0_{1\times{9}}} & 0 & 0
		\end{array}\right]$ \\
  \hline
    $n=y$
    & $\alpha =  -tr(\mathbf{J}_3), \beta = 0$ & $\alpha =  -tr(\mathbf{J}_3), \beta = 0$  &
    \multicolumn{1}{c}{
        \begin{tabular}[c]{@{}c@{}}
            \\$\mathbf{U} = \mathbf{0}_{9\times9}, \mathbf{V}= \mathbf{E}_{ij} \otimes [\mathbf{j}_3]_\times$ \\ 
            $\mathbf{u} = \mathbf{e}_k\otimes\mathbf{j}_{3}, \mathbf{v} = \mathbf{0}_{9\times1}$ \\ \\
    \end{tabular}}  \\
  \hline
    $n=\Delta \tau$
    & $\alpha = 0, \beta = -tr(\mathbf{J}_3)$ &  $\alpha = 0, \beta = -tr(\mathbf{J}_3)$ & 
    \multicolumn{1}{c}{
        \begin{tabular}[c]{@{}c@{}}
            \\$\mathbf{U} = \mathbf{0}_{9\times9}, \mathbf{V}= \mathbf{E}_{ij} \otimes [\mathbf{j}_3]_\times$ \\ 
            $\mathbf{u} = \mathbf{0}_{9\times1}, \mathbf{v} = \mathbf{e}_k\otimes\mathbf{j}_{3}$ \\ \\
    \end{tabular}}  \\
    \hline
\end{tabular}
\begin{tablenotes}
    \item Table.I: Generalized rotation matrix constraints and their corresponding constraint matrix $\mathbf{Q}_i$ for $n = y$ or $\Delta \tau$. $\mathbf{J}_3$ and $\mathbf{j}_3$ are matrix and vector will all element 1.
    $\mathbf{E}_{ij}$ and $\mathbf{e}_{i}$ are $(i,j)$th coordinate matrix and $i$th coordinate vector.
    $[\mathbf{v}]_\times$ denotes the skew symmetric matrix of a vector $\mathbf{v}$.
\end{tablenotes}
\end{threeparttable}
\vspace{-0.6cm}
\end{table*}
In this section, we solve the formulated problem to recover the time offset and relative poses simultaneously.
First, we introduce auxiliary variables and transform the original problem into a QCQP to handle the rotation constraints. 
Then, we apply semi-definite relaxation to solve the problem with global optimality.
Last, we proposed an iterative algorithm, which recovers the time offset from coarse to fine.

\subsection{QCQP Formulation}
This section aims to handle the rotation constraint involved in $\mathbf{r}_p$ and  $\mathbf{r}_s$.
Previous work \cite{wang2022certifiably} has pointed out the  generalized rotation matrix constraints for scaled rotation $n\mathbf{R}$ consist of the following three types of constraints:
\begin{gather}
	(n\mathbf{R}^{\mathbf{T}} 
 (n\mathbf{R}) = n^2\mathbf{I}_3, \label{constrain1}\\
	(n\mathbf{R})
 (n\mathbf{R})^{\mathbf{T}} =  n^2\mathbf{I}_3, \label{constrain2}\\
	(n\mathbf{R})^{(i)} \times (n\mathbf{R})^{(j)} =n(n\mathbf{R})^{(k)} \label{constrain3}, \\
	 \forall (i,j,k) = {(1,2,3),(2,3,1),(3,1,2)}, \notag 
 \end{gather}
The superscript $(\mathbf{M})^{(i)}$ denotes the $i$th column of matrix $\mathbf{M})$.
In this paper, our faced rotation matrix constraints are corresponding to $n = y$ and $n = \Delta \tau$, respectively.
Briales\cite{briales2017convex, briales2018certifiably} indicates that all above constraints can be expressed in quadratic terms, which provides an approach to transform the matrix constraint into quadratic constraints.
However, our original variable $\widetilde{\mathbf{x}}$ does not include the time offset variable $\Delta \tau$, which means we can not construct constrain using $\widetilde{\mathbf{x}}$ directly.
Therefore, we introduce an auxiliary variable $\Delta \tau$ to obtain an lifted optimization variable $z = [\ \mathbf{r}^T_s,\ \mathbf{r}^T_p,\ y,\ \Delta\tau]^T \in \mathbb{R}^{20}$.
With $z$, we can express the above constraints as 
\begin{equation}
	\begin{aligned}
           {\mathbf{z}}^T \mathbf{Q}_i {\mathbf{z}} = 0.			
	\end{aligned}
\end{equation}
The detailed structure of $\mathbf{Q}_i$ for $n=y$ and $n=\Delta \tau$ is listed in Table I.
Besides the rotation matrix constraint, the involved constraints include $y^2 = 1$ and $(\Delta\tau)(\mathbf{r}_p) = (y)(\mathbf{r}_s)$, which can also be expressed as quadratic constraints in term as $\mathbf{z}$.
Therefore, by lifting variable dimension, we transform the original problem with matrix constraint into a QCQP:
\begin{gather}
    \mathbf{z}^* =\underset{\mathbf{z}}{\min}\ {\mathbf{z}}^T \mathbf{Q_0} \mathbf{z} \notag \\
    s.t. \quad {\mathbf{z}}^T \mathbf{Q}_i {\mathbf{z}} = g_i, i=1,..,m.
\end{gather}
where $\mathbf{Q_0} = \begin{bmatrix} \bar{\mathbf{Q}} & 0 \\
		0 & 0\end{bmatrix} \in \mathbb{R}^{20 \times 20}$ and $m$ is the number of involved constraints.

\subsection{SDR and Iterative Optimization}
\label{subsec:iterative}
Since the convex programming can be solved in polynomial time and has strict global optimality,
it is appealing to transform the QCQP into a convex problem losslessly.
Due to $\mathbf{v}^T \mathbf{M} \mathbf{v} = tr(\mathbf{M} \mathbf{v} \mathbf{v}^T)$, we replace $\mathbf{z}^T\mathbf{z}$ with $\mathbf{Z}$ and drop the nonconvex rank constraint by Shor's relaxation, resulting the relaxed problem:
\begin{gather}
 \underset{\mathbf{Z}}{\min}\ tr(\mathbf{Q_0} \mathbf{Z})  \notag\\
s.t. \ \mathbf{Z} \succeq 0, tr(\mathbf{Q_i Z}) = g_i, i=1,...,m. \notag 
\end{gather}
If the solution of this problem $\mathbf{Z}^*$ meets the condition $\text{randk}(\mathbf{Z}^*) = 1$, it means the relaxation is tight, and we can exactly recover the global minimizer of the original problem.  
The tightness is widely proven in robotics when the noise upon the data matrix is limited, and our subsequent experiment also verifies it.
Therefore, given $\mathbf{Z}^*$, we can perform rank-1 decomposition to obtain $\mathbf{z}^*$ and directly recover time offset and relative poses from it.

Up to this point, we have proposed a complete method to achieve mutual localization with time offset estimation, including a novel error model, problem formulation, and lossless relaxation.
However, the entire method builds on the constant velocity assumption, which is only valid when the time offset is limited.
When the time drift between robots is sufficiently large, which is common in practice, the assumption may be violated and lead to incorrect estimation.

To address this issue, we propose an iterative method to recover the time offset from coarse to fine.
After each completed estimate, we shift the timestamps of robot $\mathcal{A}_2$'s local odometry using the estimated time offset.
Then the shifted odometry is used for the next estimation until convergence.
By this iterative method, summarized in Algorithm 1, the range of time offset to be recovered should be enhanced.
An example of the entire iteration process is illustrated in Fig.\ref{fig: head}, where $\mathcal{A}_1$ move along "IC" and $\mathcal{A}_2$ move with "RA".
After several iterations, our method recovers the correct relative pose, even when the time offset is 3s. 

\begin{algorithm}
    \caption{Iterative Time Optimization}
    \label{alg:Iter Time}
    \KwIn{ 
        Bearings from $\mathcal{A}_1$ to $\mathcal{A}_2$: $bearing_1$, \\
        Iteration times: $N_{iter}$, \\
        Robot $\mathcal{A}_1$'s local odometry: $odom_1$, \\
        Robot $\mathcal{A}_2$'s local odometry: $odom_2$}
    \KwOut{Total time offset $T$} 
    \Begin
    {   
        $T\leftarrow 0$ \\
        \For {\textbf{each} $i\in N_{iter}$ }
        {
            $\Delta \tau_i \leftarrow \text{TimeOpt}(bearing_1, odom_1, odom_2)$ \\
            $T \leftarrow T + \Delta \tau_i$ \\
            \If{$\Delta \tau_i < \epsilon$} {\textbf{break}}
            $odom_2 \leftarrow \text{ShiftOdom}(odom_2, \Delta \tau_i)$ \\
        }
        \textbf{return} $T$
    }
\end{algorithm}
\vspace{-0.4cm}

\section{Experiments}
\label{sec:experiment}

In this section, we conduct extensive experiments to evaluate the performance of our proposed method, including the \textbf{n}on-iterative \textbf{t}ime \textbf{o}ptimization (NTO) method and \textbf{i}terative \textbf{t}ime \textbf{o}ptimization (ITO) method.
In simulation experiments, the result shows that ITO has a better tolerance to time offset than NTO.
Furthermore, by comparing the two methods with the method in Wang.\cite{wang2022certifiably}, which lacks time offset estimation, our method presents better robustness and accuracy in general noisy cases.
Note that we eliminate the anonymity consideration in Wang.\cite{wang2022certifiably} for a fair comparison.
We then apply our methods in the real world to verify their practicality and robustness.
All methods are implemented in MATLAB using cvx\cite{grant2014cvx} as an SDP solver and run on a PC (Intel i5-9400F) with a single core.

\subsection{Simulation Experiment}
\label{subsec:simu exp}

To simulate data containing both poses and velocities, we follow the method in \cite{geneva2020openvins} to construct third-order SE3 B-splines for trajectory representation. 
For each robot, the corresponding B-spline is generated from 40 randomly sampled control points.
To guarantee the smoothness, the distance between two control points does not exceed 0.5m and the time interval of B-spline is set to 1 second.
We then generate synthetic data using these B-splines.
Firstly, we sample 4000 poses at 5ms intervals as the robot's local odometry. 
Then, we produce 200 bearing measurements for each pair of robots.
To control noise levels, we add Gaussian noise into bearing measurements with different standard deviations $\sigma$

\begin{figure}[t]
 \centering
 \includegraphics[width=1\columnwidth]{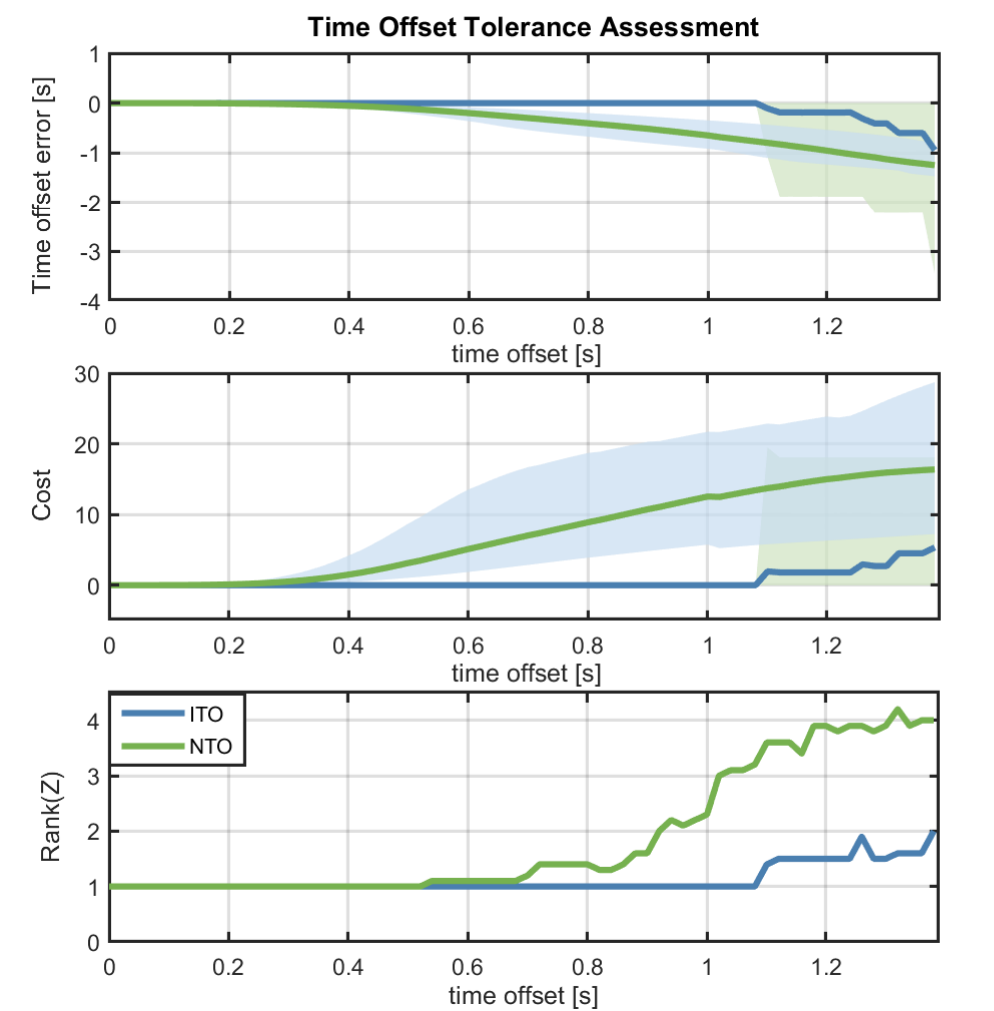}
 	\caption{\label{fig:all_error} The change of time offset estimation error, objective cost, and rank of solution $\mathbf{Z}^*$ under different time offsets between robots. }
 	\vspace{-0.5cm}
\end{figure}

\subsubsection{Tolerance and Optimality Comparison}
In this experiment, we aim to evaluate the influence of iterative optimization in time offset estimation in both tolerance level and optimality.
We add no noise in bearing measurements and change the time offset of two robots' trajectories on purpose from 0s to 1.4s with an interval of 0.02s.
 We perform 100 Monte-Carlo simulations with randomly generated B-spline for each determined time offset.
 The time offset estimation, cost, and the rank of the solution are shown in Fig.\ref{fig:all_error}.
 It shows the NTO can not recover the correct time offset when it exceeds around 0.4s.
 The reason is that the uniform velocity assumption can not hold when the time offset is too large. 
 In contrast, the ITO method can maintain optimality even for time offsets exceeding 1s and still precisely estimate the time offset.
 It suggests that the from-coarse-to-fine algorithm effectively extends the applicability of the uniform velocity assumption in extreme situations.
 In summary, ITO outperforms NTO in terms of time tolerance and optimality, especially in more challenging scenarios (large time offset), which validates our perspective in Sec.\ref{subsec:iterative}.

\subsubsection{Estimation Accuracy Comparison}
\begin{figure} [t]
 \centering
 \includegraphics[width=1\columnwidth]{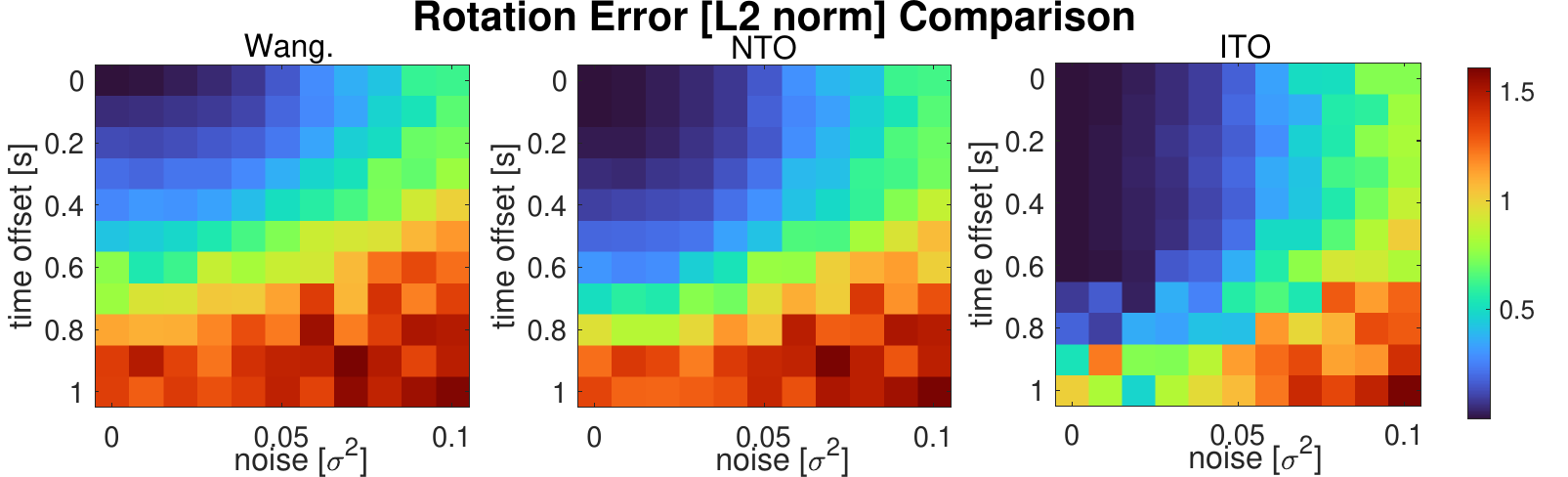}
 	\caption{\label{fig:rot_error} Distribution of relative rotation estimation error with change of time drift and noise level.}
 	\vspace{-0.0cm}
\end{figure}

\begin{figure}  [t]
\centering
 \includegraphics[width=1\columnwidth]{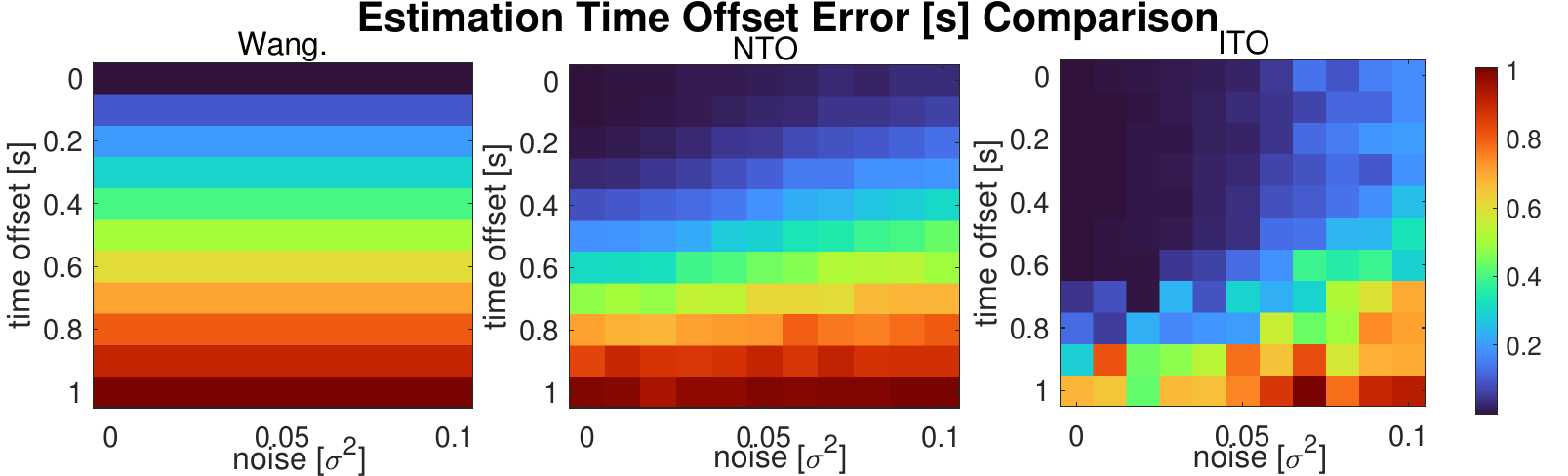}
 	\caption{\label{fig:time_offser_error} Distribution of time offset estimation error with change of time drift and noise level.}
\vspace{-2.0cm}
\end{figure}

In this experiment, we aim to evaluate our method in more general cases for comprehensive assessments.
 We compare our methods (NTO and ITO) with Wang.\cite{wang2022certifiably} in two dimensions, times offset and noise level.
 We change the times offset (from 0 to 1s with 0.1s interval) and noise (from 0 to 0.1 with 0.01 interval) respectively, and perform 100 trials for each configuration, resulting in 11000 data.
 Two metrics, the mean estimation error and the meantime offset estimation error, are considered in this experiment.
 The results are shown in Fig.\ref{fig:rot_error} and Fig.\ref{fig:time_offser_error}, respectively.
 Note that since there is no time offset estimation in Wang.\cite{wang2022certifiably}, we take zeros as its estimation result to compute the error. 
 
 We obtained two findings from the results of these experiments.
 First, as the time offset and noise level change, the error distributions of both pose estimation and time offset estimation show a clear consistency.
 It means better time offset estimation can improve mutual localization's performance, highlighting the significance of time synchronization.  
 Second, the results show that the dark region of ITO is larger than NTO, while that of Wang\cite{wang2022certifiably} is the smallest, especially in areas where time offset is large ($>0.5$).
 It means both NTO and ITO outperform the existing method, with ITO emerging as the superior algorithm among the three algorithms.
 This experiment presents how our proposed method extends the applicability of mutual localization in time-drifted and noisy cases.

\subsection{Real-world Experiments}

\begin{figure}  [b]
\centering
\vspace{-0.1cm}
 \includegraphics[width=1\columnwidth]{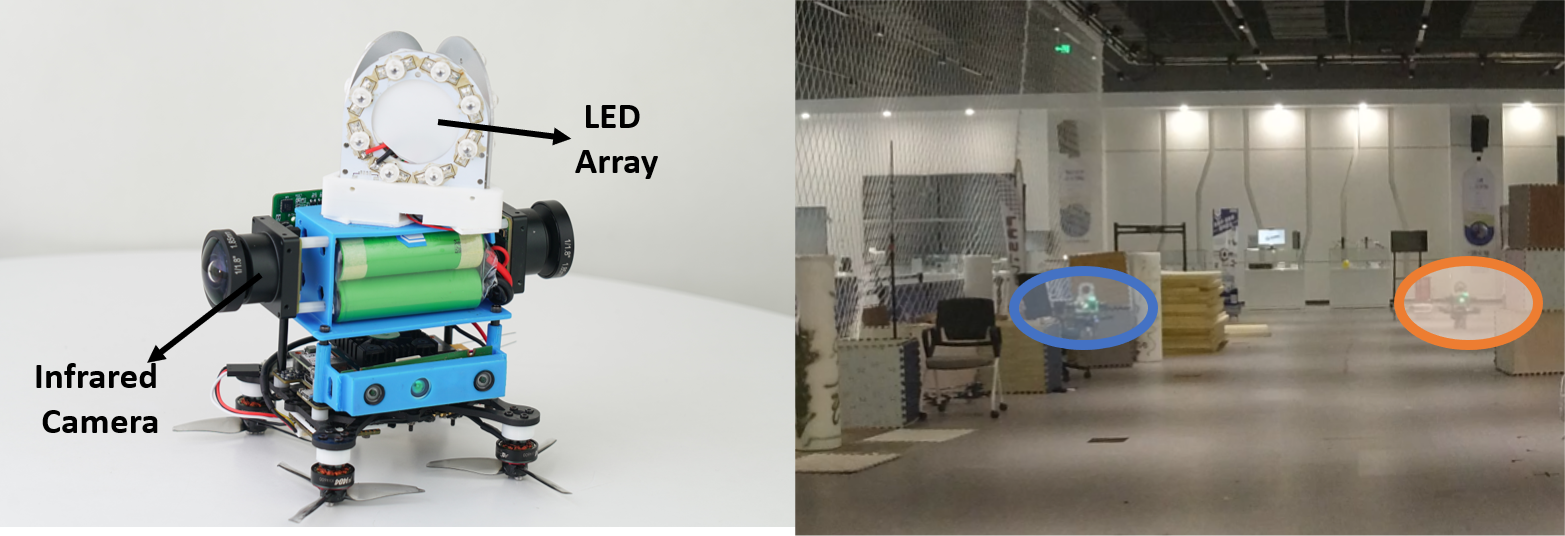}
 	\caption{\label{fig:real_exp} Platform and snapshot in real-world experiment.}
\vspace{-0.4cm}
\end{figure}

In this part, we apply our method to real-world data to evaluate the practicality and robustness.
 The quadrotor platforms and a snapshot of our experiments are shown in Fig.\ref{fig:real_exp}.
 Quadrotors are equipped with LED arrays and infrared cameras for mutual detection.
 We allow robots to move along pre-determined 3D trajectories and obtain stable inter-robot bearing measurements during flight.
 We conducted totally five experiments with different trajectory patterns, including pentagons, figure-eight, and squares.
 In each experiment, we fix the time offset between trajectories as 1s and utilize motion capture as the ground truth.

\begin{figure} [t]
 	\centering
 	\includegraphics[width=1\columnwidth]{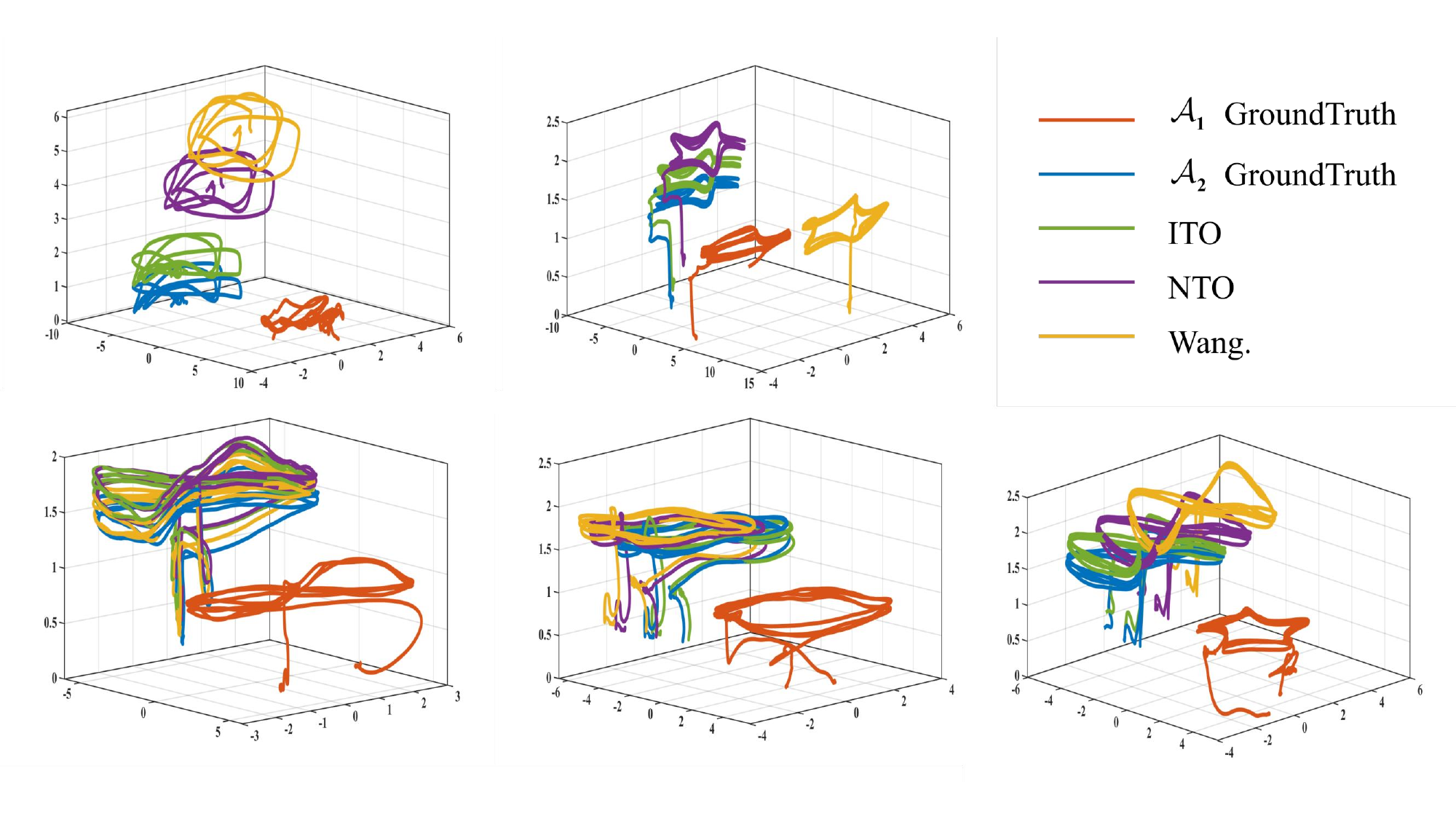}
 	\caption{\label{fig:real_traj} Comparisons of estimated trajectories in real-world experiments. Two robots' trajectories are shown in a common frame.}
 	\vspace{-0.0cm}
 \end{figure}

 With collected data, we compare the above-mentioned three methods in multiple dimensions: optimization cost, estimation error, estimated time offset, and runtime.
 As Table.II shows Wang.\cite{wang2022certifiably} always converge to a larger cost and obtain egregiously
large error in translation and rotation. 
    The erroneous estimation comes from the incorrect correspondence between bearing measurement and observed robot's local odometry. 
 Compared with other methods, ITO consistently obtains the minimum cost in all experiments with the most or second most small estimation error. 
 Note that, due to noise, obtaining the minimum cost does not mean obtaining the most accurate estimation.
 In the aspect of time offset estimation, ITO consistently recovers the accurate time offset while NTO has a significant error.
 It is clear that ITO needs more computation due to iteration.
 But it is acceptable since time synchronization can run as a background program periodically in multi-robot systems. 
 The aligned trajectories using estimated relative pose by three methods are illustrated in Fig.\ref{fig:real_traj}.
 It shows ITO can still work properly to recover a correct relative pose even with time-drifted odometry data, while other methods fail. 
 These experiments verify the practicality and robustness of our proposed method.

\begin{table}[t]
\centering
\caption{Results of real-world experiments.}
\begin{tabular}{cccccc} 
\hline
\multicolumn{1}{l}{Method} & \multicolumn{1}{l}{Cost} & Trans(m)       & Rot($\degree$)           & \begin{tabular}[c]{@{}c@{}}Estimation \\Time Offset(s)\end{tabular} & \multicolumn{1}{l}{Runtime(s)}  \\ 
\hline
\multirow{5}{*}{Wang.\cite{wang2022certifiably}}       & 65.524                   & 2.82           & 28.26         & -                                                                & 0.418                         \\
                           & 51.776                   & 6.52           & 144.19        & -                                                                & 0.441                         \\
                           & 18.184                   & \textbf{0.17}  & \textbf{1.16} & -                                                                & 0.436                         \\
                           & 15.516                   & 0.72           & 10.78         & -                                                                & 0.465                         \\
                           & 92.19                    & 1.11           & 21.57         & -                                                                & 0.465                         \\ 
\hline
\multirow{5}{*}{NTO}       & 47.409                   & 1.834          & 18.34         & 0.374                                                            & 0.408                         \\
                           & 30.896                   & 0.448          & 27.43         & 0.659                                                            & 0.438                         \\
                           & 5.305                    & 0.187          & 1.85          & 0.698                                                            & 0.437                         \\
                           & 10.492                   & 0.536          & 7.29          & 0.377                                                            & 0.432                         \\
                           & 57.251                   & 0.514          & 12            & 0.545                                                            & 0.464                         \\ 
\hline
\multirow{5}{*}{ITO}       & \textbf{3.994}           & \textbf{0.489} & \textbf{4.4}  & 0.905                                                            & 2.064                         \\
                           & \textbf{5.393}           & \textbf{0.163} & \textbf{3.39} & 0.962                                                            & 1.687                         \\
                           & \textbf{1.799}           & 0.202          & 1.53          & 0.963                                                            & 1.75                          \\
                           & \textbf{2.268}           & \textbf{0.049} & \textbf{4.01} & 0.869                                                            & 2.177                         \\
                           & \textbf{4.22}            & \textbf{0.142} & \textbf{2.87} & 0.961                                                            & 2.337                         \\
\hline
\end{tabular}
\vspace{-0.5cm}
\end{table}

\section{Conclusions and Future Work}
\label{sec:conclusion}
In this paper, we proposed a complete framework to simultaneously achieve time synchronization and mutual localization.
The joint estimation of time offset and relative poses significantly improves the performance of mutual localization when there is a certain time drift between robots.
Extensive experiments on simulated and real-world data show the outperforming accuracy of our method compared with existing methods.
In the future, we will apply our mutual localization method in various multi-robot applications to facilitate efficient collaboration among multiple robots.
\bibliography{ICRA}
\end{document}